\documentclass[sigconf]{acmart}

\usepackage{subcaption} 
\AtBeginDocument{%
  }

\setcopyright{acmlicensed}
\copyrightyear{2024}
\acmYear{2024}

\acmConference[ACM KDD 2024 | Barcelona, Spain]{}{August 25-29,
  2024}

\acmISBN{978-x-xxxx-xxxx-x/24/08}

\begin{document}

\title{A COMPARATIVE STUDY OF DEEP REINFORCEMENT LEARNING MODELS: DQN VS PPO VS A2C}

\author{Neil de la Fuente}
\authornote{Both authors contributed equally to this research.}
\email{Neil.DeLaFuente@autonoma.cat}
\affiliation{%
  \institution{Autonomous University of Barcelona}
  \city{Barcelona}
  \country{Spain}
}
\affiliation{%
  \institution{Computer Vision Center}
  \city{Barcelona}
  \country{Spain}
}

\author{Daniel A. Vidal Guerra}
\authornotemark[1]
\email{DanielAlejandro.Vidal@autonoma.cat}
\affiliation{%
  \institution{Autonomous University of Barcelona}
  \city{Barcelona}
  \country{Spain}
}

\affiliation{%
  \institution{Computer Vision Center}
  \city{Barcelona}
  \country{Spain}
}


\renewcommand{\shortauthors}{Neil de la Fuente and Daniel Vidal}

\begin{abstract}
This study conducts a comparative analysis of three advanced Deep Reinforcement Learning models – Deep Q-Networks (DQN), Proximal Policy Optimization (PPO), and Advantage Actor-Critic (A2C) – exclusively within the BreakOut Atari game environment. Our research aims to assess the performance and effectiveness of these models in a singular, controlled setting. Through rigorous experimentation, we examine each model's learning efficiency, strategy development, and adaptability under the game's dynamic conditions. The findings provide critical insights into the practical applications of these models in game-based learning environments and contribute to the broader understanding of their capabilities in specific, focused scenarios. Code is publicly available: \href{https://github.com/Neilus03/DRL_comparative_study}{github.com/Neilus03/DRL\_comparative\_study}
\end{abstract}

\keywords{Deep Reinforcement Learning, BreakOut Atari Game, DQN, PPO, A2C}

\maketitle

\section{Introduction}
In this comprehensive study, we explore the complex domain of Deep Reinforcement Learning (DRL), focusing on a thorough comparison of three well-known models: Deep Q-Networks (DQN), Proximal Policy Optimization (PPO), and Advantage Actor-Critic (A2C), specifically within the BreakOut Atari game environment. To ensure consistency and robustness in our experiments, we exclusively utilized the well-established implementations of DQN, PPO, and A2C from the Stable Baselines3 (SB3) framework \cite{raffin2021}. This methodology provides a clear and equitable platform for comparing the nuances of each model's learning strategy, adaptability, and efficiency within a controlled environment.

Our methodology included a detailed exploration of the hyperparameter settings for each model to comprehend their impact on performance. Specifically, we varied the learning rates across the DQN, PPO, and A2C models from SB3 to identify how these rates influenced the speed and efficiency of learning, as well as the overall strategy development within the game, and gamma discount factors to examine how short-term versus long-term reward prioritization affected the models' decision-making processes and overall performance.

This aspect of our study was key for revealing each model's capability to balance immediate rewards with future ones, a critical consideration in many real-world applications of DRL. Through this comprehensive approach, our research provides insights into the optimal configuration of these models for efficient and effective learning in complex environments.

\section{Related Work}
Deep Reinforcement Learning has emerged as a significant field combining Deep Learning (DL) with Reinforcement Learning (RL), notably impacting areas like gaming and robotics. The development of DQNs \cite{mnih2013} marked a turning point in DRL by demonstrating the effective use of neural networks with Q-learning to play Atari 2600 video games. It set a new precedent for applying DL to approximate Q-values. Subsequent enhancements, such as Double DQN \cite{van2016} and Dueling DQN \cite{wang2016}, addressed key issues like Q-value overestimation and improved DQN's stability.

PPO, introduced in \cite{schulman2017}, advanced policy gradient methods. PPO is especially noted for its efficient balance between sample efficiency and ease of implementation. Its stable and reliable policy updates have made it one of the preferred choice in the DRL community.

A2C, a simplified version of the Asynchronous Advantage Actor-Critic (A3C), has been influential in actor-critic methods. A2C, while maintaining the dual advantages of learning policy and value functions, removes the complexity of asynchronous operations, as described in \cite{mnih2016}.

Comparative studies in gaming environments, such as the one conducted in \cite{henderson2018}, provide insight into the practical applications of these models and help to understand their strengths and weaknesses. However, challenges like sample efficiency, stability, and generalization still persist in DRL. Current research efforts in this field are focused on addressing these issues to enhance the efficiency, scalability, and broader applicability of DRL models.

\section{Methodology}
Our experiments focused on the influence of varying hyperparameters. We tested four to five learning rates for each model to observe the effects on convergence and training robustness, and gamma values to explore the prioritization of long-term versus immediate rewards. The training for each model was consistently conducted over a fixed number of episodes, with episode reward and learning stability as our primary performance metrics. This approach facilitated a controlled and replicable comparison across the models, ensuring that any observed differences in performance were attributable to the models' intrinsic characteristics and the selected hyperparameters.

\subsection{Implementation of the models}
Opting for SB3 allowed us to ensure that each of the three models adhered to the highest standards and best practices within the field of DRL.

The DQN implementation is directly inspired by the architecture outlined in \cite{mnih2013}. The consistent and reliable performance of the SB3 version of DQN offered a robust foundation for our comparative exploration, particularly within the context of the BreakOut game. PPO is particularly notable for its balance between sample efficiency and simplicity in implementation, aligning with the original design principles laid out by \cite{schulman2017}. Finally, the A2C model maintains the core advantages of A3C while removing the need for asynchronous operations, as detailed in \cite{mnih2016}.

\subsection{Hyperparameter Variations}
We explored a spectrum of learning rates to evaluate the different speeds of learning and robustness of strategy development. In tandem, we manipulated the gamma discount factor with two distinct values, 0.99 and 0.90, to assess the models' sensitivity to short-term versus long-term rewards.

Our structured approach provides a foundation for an in-depth evaluation of the learning dynamics and performance efficacy of each model, which we will further develop in the 'Experiment Setup' and 'Results' sections.

\section{Hypotheses and Theoretical Considerations}
We formed several hypotheses based on the theoretical foundations of the models in question and their known strengths and weaknesses. Here we reflect both the predictive and the conceptual nature of the content. In the following, we describe our hypotheses for the performance of the different models implemented.

\subsection{Alignment of BreakOut Dynamics with DQN's Value Estimation}
In the BreakOut game, the dynamics are particularly well suited for the Q-learning algorithm, upon which the DQN is based. Q-learning seeks to learn a value function \( Q(s, a) \), representing the expected return of taking an action \( a \) in a state \( s \) and following the optimal policy thereafter. For BreakOut, the state \( s \) can be described by the position of the paddle, the ball, and the configuration of the bricks, while the action \( a \) corresponds to moving the paddle left, right, or staying in place.

The game's immediate and clear reward structure—points gained from breaking bricks—aligns with the value estimation approach of DQN. The Q-function for DQN is updated using the following rule:

\[
\resizebox{\columnwidth}{!}{
$Q_{\text{new}}(s_t, a_t) \leftarrow Q(s_t, a_t) + \alpha \left[r_{t+1} + \gamma \max_{a'} Q(s_{t+1}, a') - Q(s_t, a_t)\right]$
}
\]

where \( \alpha \) is the learning rate, \( \gamma \) is the discount factor, \( r_{t+1} \) is the immediate reward, and \( \max_{a'} Q(s_{t+1}, a') \) is the maximum predicted value for the next state \( s_{t+1} \), over all possible actions \( a' \). This formula allows DQN to incrementally improve its policy by learning from the immediate outcomes of its actions, which in BreakOut are both straightforward and tightly coupled with the action taken.

Contrastingly, on-policy methods like PPO and A2C involve estimating the policy directly. The policy \( \pi(a|s) \) represents the probability of taking action \( a \) in state \( s \), and is adjusted in the direction suggested by the policy gradient:
\[
\nabla_{\theta} J(\pi_\theta) = \mathbb{E}_{\tau \sim \pi_\theta} \left[ \sum_{t=0}^{T} \nabla_{\theta} \log \pi_\theta(a_t | s_t) \cdot A^{\pi_\theta}(s_t, a_t) \right]
\]

The advantage function \( A^{\pi_\theta}(s_t, a_t) \), which measures the relative value of an action compared to the average, is more complex to estimate in a game like BreakOut. The reason is that certain strategies, like "tunneling", where the ball is directed behind the brick wall to clear multiple bricks, require careful planning and a detailed understanding of the game's physics, which are not immediately obvious from the reward signal.

Estimating the value of strategies can be more challenging than directly mapping actions to outcomes, as done in Q-learning. DQN's value estimation through Q-learning is more straightforward because it can directly correlate actions like "move the paddle under the ball" with receiving points for breaking bricks. In contrast, PPO and A2C must learn through trial and error which complex sequences of actions yield higher returns. This makes DQN potentially more suited for BreakOut, where the optimal policy involves clear and direct action-reward correlations, making the value function approach more effective than direct policy estimation in this context.

\subsection{Advantage of Experience Replay in DQN}
The concept of experience replay is fundamental to the success of DQN. In the BreakOut game, the mechanics often lead to repetitive scenarios where the paddle must hit the ball in a rhythmic pattern. These recurrent interactions between the ball and paddle could potentially lead to a form of learning that is too specialized or optimized for the frequent but limited scenarios encountered.

The experience replay mechanism mitigates this risk by storing a history of experiences: \[ e_t = (s_t, a_t, r_{t+1}, s_{t+1})\]

in a replay buffer \( D_t \). This buffer acts as a diversified data reservoir from which the DQN samples to update the value function.

This process ensures that the DQN does not simply memorize the most recent or frequent patterns but instead develops a more generalized understanding of the game by learning from a broader range of experiences.

The advantage of such an approach in BreakOut is twofold: it helps DQN avoid local optima by not getting stuck in repetitive but suboptimal strategies, and it allows DQN to learn from rare, high-reward episodes, such as when the ball reaches the back of the brick wall creating a 'tunnel', ensuring these crucial experiences are not lost.

Continuously revisiting diverse past experiences makes DQN's training more robust and less prone to overfitting. This is especially beneficial in BreakOut, where subtle differences between good and great strategies are crucial. Experience replay enhances both learning and retention of effective strategies, which is essential for mastering repetitive and pattern-based environments.

\subsection{Sensitivity to Hyperparameters Across Models}
The sensitivity of an algorithm to its hyperparameters can greatly influence its performance. Our hypothesis suggests that DQN and PPO will be less sensitive to learning rate variations, while A2C will display greater sensitivity.

DQN's use of experience replay and fixed Q-targets provides a stabilizing effect on learning updates. The experience replay randomizes the data, thus breaking correlations in the observation sequence and smoothing changes in the data distribution. This leads to stable gradient descent updates, allowing DQN to handle a wider range of learning rates.

The robustness of PPO comes from its objective function, which uses a clipping mechanism to prevent excessively large policy updates that could lead to instability. This clipping mechanism acts as a safeguard against the potential negative effects of choosing a suboptimal learning rate.

On the other hand, A2C's performance is closely tied to its learning rate. As an on-policy algorithm, A2C continuously updates its policy based on the latest data it collects from the environment. If the learning rate is too low, A2C's policy may not adapt quickly enough to new information, leading to suboptimal performance. On the contrary, a high learning rate might cause the policy to change too fast, destabilizing the learning process. Therefore, the ideal learning rate for A2C should balance quick data integration and instability avoidance.

Discount factors also play a crucial role in the learning process. A high discount factor \( \gamma \) makes an algorithm more farsighted by placing more emphasis on future rewards. This could potentially be more beneficial for DQN, as it may enable the agent to better recognize the long-term benefits of strategic moves like tunneling. A lower \( \gamma \) could make the agent myopic, prioritizing immediate rewards, which might suffice for simpler strategies but fail to capture the depth of the game's strategic possibilities.

In the case of PPO and A2C, a well-chosen discount factor helps in balancing the trade-off between exploring new strategies and exploiting known rewarding behaviors. PPO's ability to maintain stable updates might be less affected by the discount factor compared to A2C, as the latter can be more sensitive to the choice of \( \gamma \) due to its direct policy update mechanism.

In summary, we hypothesize that DQN and PPO will exhibit a higher tolerance to hyperparameter variations, while A2C will require careful tuning of its learning rate and discount factor to achieve optimal performance. This sensitivity to hyperparameters is an essential consideration in the deployment and tuning of reinforcement learning models in practice.

\subsection{Policy Optimization Stability in PPO and A2C}
The stability of policy optimization in PPO and A2C  depends on their distinct approaches to updating policies. PPO is designed to mitigate policy volatility through a clipping mechanism in its objective function, which constrains the extent of policy updates, fostering stability. The clipped objective in PPO is given by:

\[
\resizebox{\columnwidth}{!}{
$L^{\text{CLIP}}(\theta) = \mathbb{E}_t \left[ \min(r_t(\theta) \hat{A}_t, \text{clip}(r_t(\theta), 1-\epsilon, 1+\epsilon) \hat{A}_t) \right]$
}
\]

where \( r_t(\theta) \) is the ratio of the new policy to the old policy probabilities, \( \hat{A}_t \) is the estimator of the advantage function at time \( t \), and \( \epsilon \) is a hyperparameter that defines the clipping range.

Conversely, A2C updates policies after every step, lacking PPO's clipping barriers, which can lead to more aggressive policy changes:

\[ \Delta\theta = \alpha \nabla_\theta \log \pi_\theta(a_t | s_t) A(s_t, a_t) \]

The absence of a clipping mechanism in A2C allows for larger and more disruptive policy updates, especially when the advantage \( A(s_t, a_t) \) is significant.

We expect PPO to exhibit more stable learning progression due to its conservative update strategy, whereas A2C may show more fluctuations in its performance. This variability reflects A2C's responsiveness to immediate environmental changes, which can result in rapid but occasionally unstable learning trajectories.

These hypotheses serve not only as predictions to be tested against the empirical data but also as a framework to interpret the complexities of model behaviors. They reflect a balance between established theoretical knowledge and the particularities of BreakOut.

\section{Experimental Setup}
We tested the capabilities of DQN, PPO, and A2C by conducting our experiments within the BreakOut Atari game environment, leveraging the Gymnasium framework \cite{farama2021}.

\subsection{Parameter Variations}
\begin{itemize}
\item \textbf{Learning Rate}: We experimented with learning rates at several magnitudes: \(1e-5, 5e-5, 1e-4, 5e-4, 1e-3, 5e-3\).
\item \textbf{Gamma Discount Factor}: we varied the gamma discount factor between \(\gamma = 0.99\) and \(\gamma = 0.90\).
\end{itemize}

For our experiments, training duration was quantified in terms of frames rather than time or episodes, providing a consistent measure across all models. Each model was trained for approximately 20 million frames. This metric offers an objective standard of comparison, ensuring no model is unintentionally favored by the measurement approach.

The frame-based approach aligns with the conventions of the SB3 framework and allows for a fair assessment of each model's learning process, as it accounts for variations in episode lengths. For instance, more skilled agents may play fewer but longer episodes.

An additional consideration is the potential for the game to enter into loops, such as the ball bouncing between the wall and ceiling repeatedly without hitting bricks. Such scenarios can artificially inflate the number of frames without corresponding to meaningful learning or gameplay progress. This effect is evident when certain training runs show extended plateaus in learning curves or spikes in episode duration, which may indicate the agent has entered a loop rather than achieving a breakthrough in strategy.

By standardizing the training duration across models via frame count, we mitigate these factors, providing a clear, unbiased view of each model's learning efficiency and performance.

\subsection{Evaluation Metrics}
To assess the effectiveness of each model, we employed multiple evaluation metrics. These metrics are critical as they capture various aspects of the learning process and the models' proficiency in navigating the environment.

\textbf{Average Reward per Episode}: Measures the mean score per episode. Higher scores indicate better strategies and more efficient gameplay.

\textbf{Episodes to Threshold}: Counts the episodes needed to consistently reach reward thresholds, indicating how fast the model learns in terms of episodes.

\textbf{Time to Threshold}: Measures the hours needed to reach reward thresholds, showing learning speed.

\textbf{Reward Distribution}: Shows reward spread and variation, indicating performance consistency. Narrow spread suggests stability; wide spread suggests exploratory behaviour and instability.

\textbf{Stability of Learning}: Monitors performance fluctuations over time. Smooth progression indicates steady learning; large variations suggest instability.

\textbf{Frame Utilization Efficiency}: Evaluates how efficiently models use allocated frames, indicating purposeful learning and avoiding wasteful gameplay with loops.

Each of these metrics contributes to a composite picture of model performance. By examining these various facets, we can discern not only which model achieved the highest scores but also understand the nuances of their learning processes. This multifaceted evaluation is designed to dissect the models' responses to different configurations within the standardized BreakOut environment, setting the stage for a detailed analysis of their respective strengths and weaknesses.

\section{Results}
\begin{figure*}[htbp]
    \centering
    \begin{subfigure}[b]{7cm}
        \includegraphics[width=8cm, height=7cm]{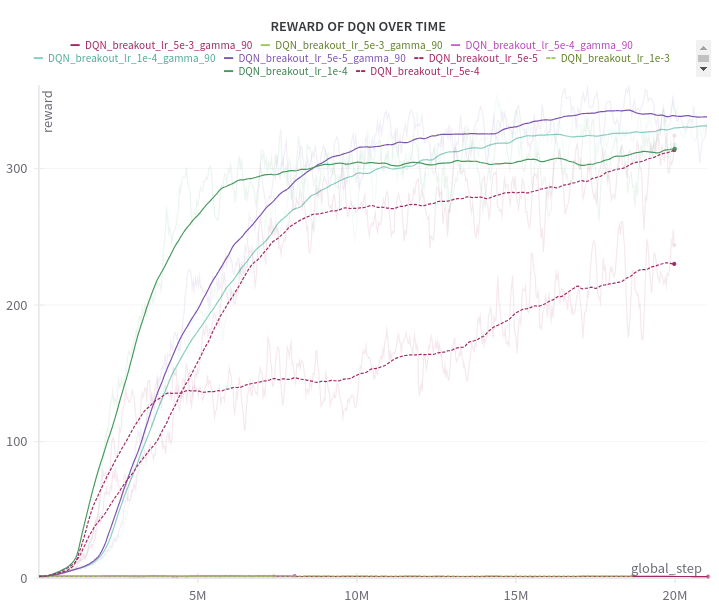}
        \caption{DQN Performance}
        \label{fig:DQN_performance}
    \end{subfigure}
    \hspace{0.9cm}
    \begin{subfigure}[b]{7cm}
        \includegraphics[width=8cm, height=7cm]{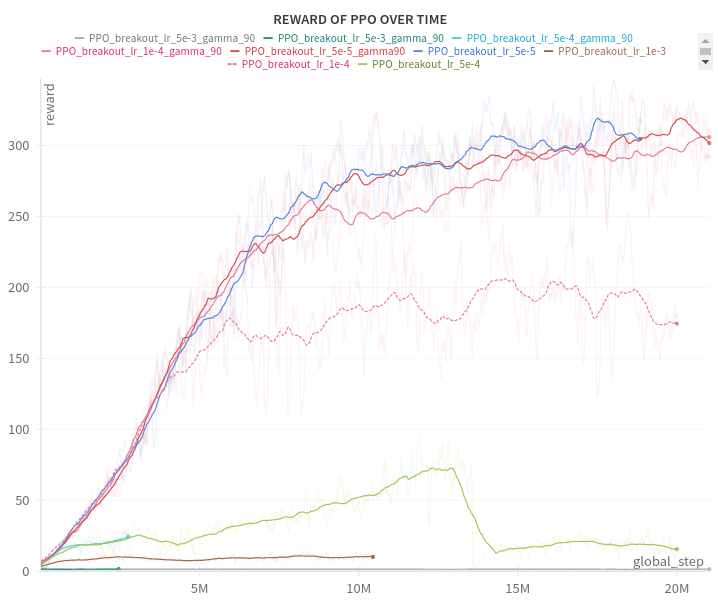}
        \caption{PPO Performance}
        \label{fig:PPO_performance}
    \end{subfigure}
    \begin{subfigure}[b]{7cm}
        \includegraphics[width=8cm, height=7cm]{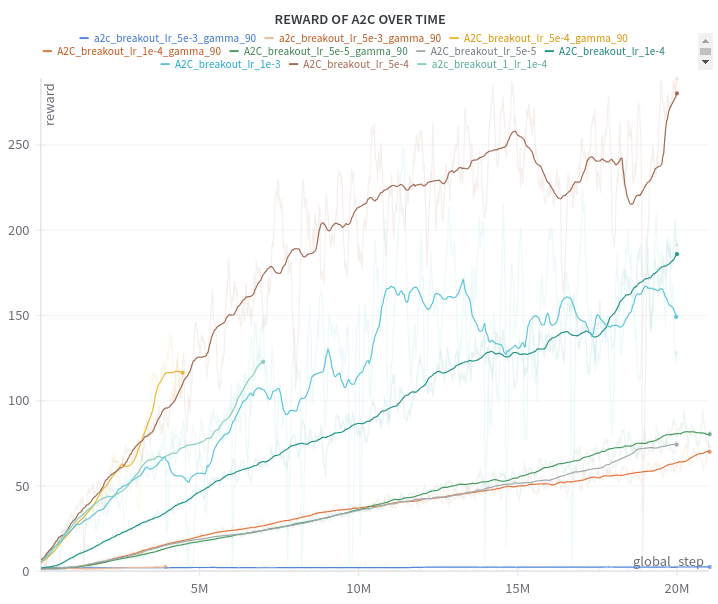}
        \caption{A2C Performance}
        \label{fig:A2C_performance}
    \end{subfigure}
    \hspace{0.9cm}
    \begin{subfigure}[b]{7cm}
        \includegraphics[width=8cm, height=7cm]{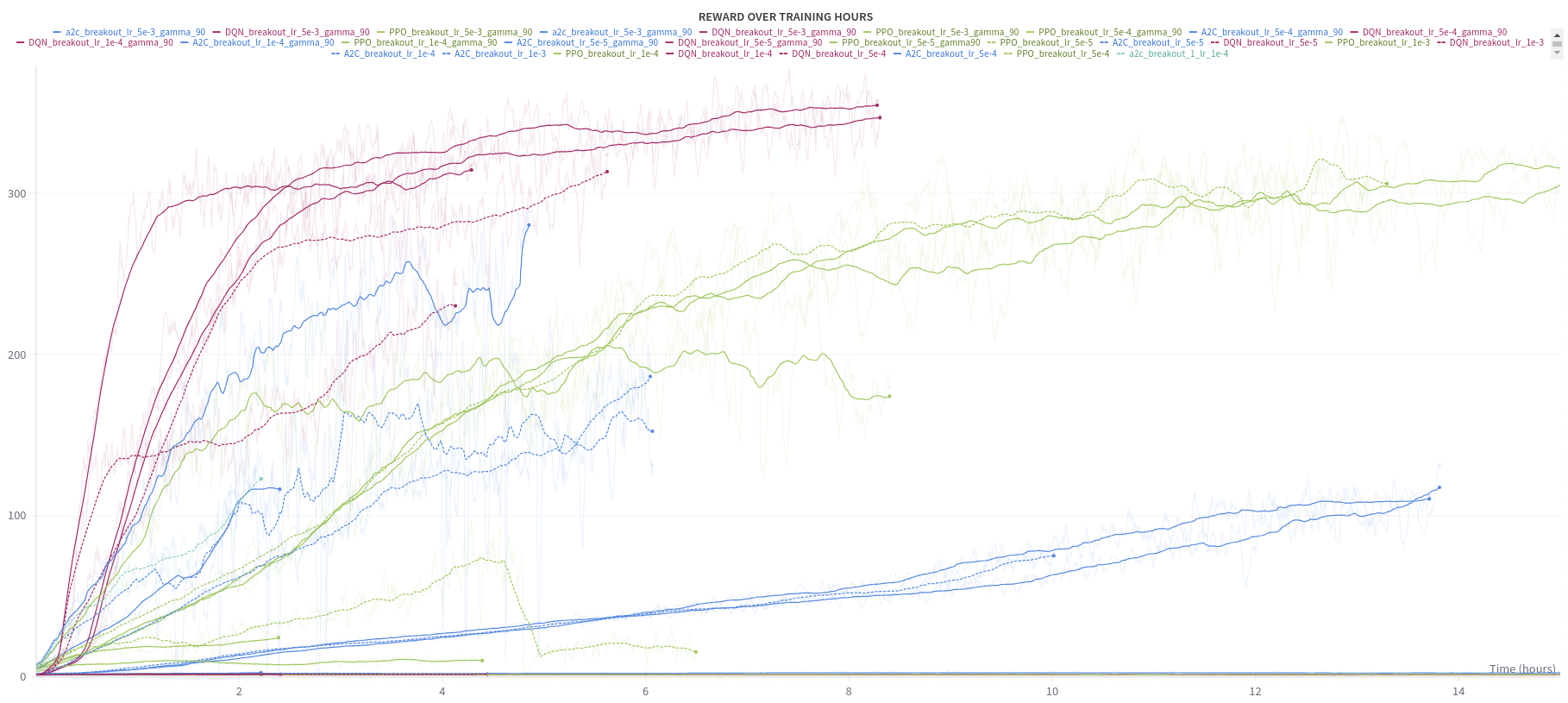}
        \caption{Reward over training hours}
        \label{fig:Reward_over_training_hours}
    \end{subfigure}
    \begin{subfigure}[b]{7cm}
        \includegraphics[width=8cm, height=7cm]{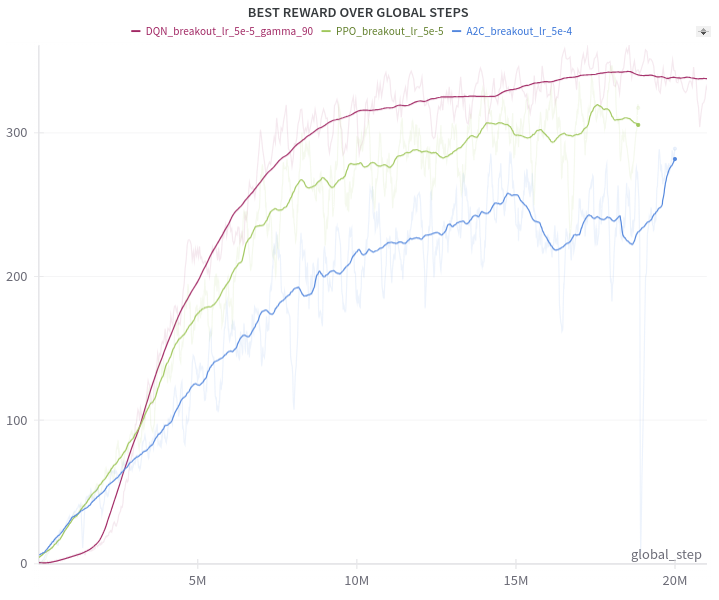}
        \caption{Reward over training steps for the best model}
        \label{fig:Reward_over_steps_best_model}
    \end{subfigure}
    \hspace{0.9cm}
    \begin{subfigure}[b]{7cm}
        \includegraphics[width=8cm, height=7cm]{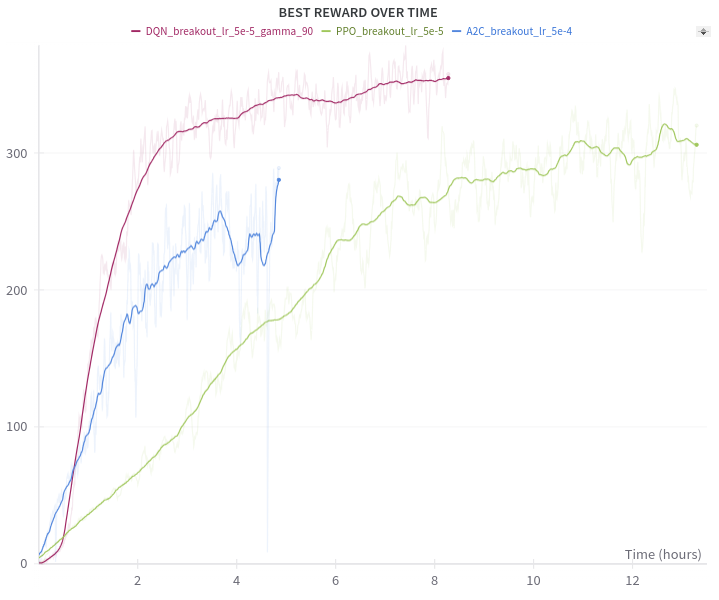}
        \caption{Reward over training time for the best model}
        \label{fig:Reward_over_time_best_model}
    \end{subfigure}
    \caption{Performance metrics for DQN, PPO, and A2C models across different settings.}
    \label{fig:performance_metrics}
\end{figure*}

Our empirical investigation into the performance of DQN, PPO, and A2C  within the BreakOut Atari game environment yielded insightful findings, as detailed in the subsections below. Figure \ref{fig:performance_metrics} showcases the performance metrics, providing a comprehensive comparison of the DQN, PPO, and A2C models across all the configurations.

\subsection{Performance Across Learning Rates}
From the data, DQN showed, in Figure ~\ref{fig:DQN_performance}, remarkable resilience across a broad span of learning rates, consistent with our hypothesis that its experience replay buffer would confer robustness against the variability in learning rates.

PPO also showed resistance to changes in learning rate as it can be seen in Figure ~\ref{fig:PPO_performance}, though its performance was more sensitive to the extremes. At the lower end of the learning rate spectrum, PPO's progress was gradual but consistent. However, at higher rates, PPO's performance was more variable, likely due to the algorithm's clipped objective function which, while mitigating the risk of large destructive updates, also caps the potential rapid advancements that can be achieved with more aggressive learning rates.

In contrast, A2C's performance was heavily influenced by the learning rate as shown in Figure ~\ref{fig:A2C_performance}. At lower rates, A2C's learning curve was almost dead, suggesting an inability to make significant policy improvements. As the learning rate increased, A2C's performance improved markedly, confirming our prediction that A2C requires a careful balance in learning rate settings to optimize its continuous policy updates.

\subsection{Adaptability to Discount Factor Variations}
The gamma ($\gamma$) discount factor influences the agent's consideration of future rewards. Across all models, a higher $\gamma$ value typically correlated with a more strategic play, where long-term gains were prioritized. DQN's performance peaked at a moderate $\gamma$ value before declining, suggesting a sweet spot where the agent was sufficiently foresighted without being impeded by the overvaluation of distant future rewards.

PPO and A2C demonstrated a clearer preference for a higher $\gamma$ value. This was particularly true for PPO, which displayed an steady increase in performance as the $\gamma$ value rose, aligning with the algorithm's inherent stability and its capability to evaluate long-term strategies effectively.

\subsection{Learning Stability, Efficiency, and Reward Optimization}
In examining the learning process of DQN, PPO, and A2C models, we considered several critical aspects: the stability and efficiency of learning, reward optimization, and episode length within the BreakOut Atari game. DQN's performance stood out, demonstrating a smooth learning curve and efficient frame utilization, highlighting its capability to rapidly assimilate and apply successful strategies. This was mirrored in its real-time training efficiency, where DQN consistently achieved high rewards in shorter durations, making it an ideal candidate for scenarios demanding quick adaptation and time-efficient learning. This can be clearly seen in Figure ~\ref{fig:Reward_over_training_hours}.

Contrastingly, PPO and A2C experienced more pronounced fluctuations in their learning trajectories. PPO, though displaying a reasonable level of stability, tended to engage in lengthier episodes, suggesting a propensity for a more exploratory approach that may sacrifice swiftness for thoroughness. A2C's learning curve was the most variable, reflecting its sensitivity to environmental dynamics and possibly a greater need for exploration to refine its policy. The higher number of frames A2C required to reach performance levels comparable to DQN implies a less efficient learning process, especially marked when considering the model's real-time performance, which lagged behind the others.

Furthermore, reward optimization and episode length analysis revealed strategic distinctions among the models. DQN's strategy excelled in securing high scores efficiently, as seen in Figures ~\ref{fig:Reward_over_steps_best_model} and ~\ref{fig:Reward_over_time_best_model}, translating learning into performance gains with remarkable speed. In contrast, PPO and A2C, despite their eventual improvements, needed longer episodes for similar levels of reward, indicating potentially less efficient strategies. This behavior underscores a strategic divergence where DQN prioritizes exploitation of the known dynamics, while PPO and A2C seem to invest heavily in exploration, an approach that may lead to substantial benefits in less structured and more complex environments, but maybe not in simpler environments like BreakOut.

The practical implications of our findings are significant. They suggest that while DQN shines in environments requiring rapid learning and deployment, PPO and A2C may offer advantages in scenarios where the depth of exploration and comprehensive strategy development are key. Such insights are highly valuable for the improved selection and application of reinforcement learning models, emphasizing the importance of aligning the model's strengths with the specific demands and constraints of the task and environment.

In summary, our results substantiate the hypotheses posed in our theoretical considerations, illustrating the nuanced interplay between model architectures, hyperparameters, and the BreakOut game dynamics. The collected data provides a comprehensive picture of how each model's unique characteristics influence its learning trajectory and overall performance in a standardized environment. These insights prompt a reevaluation of the preferred use cases for these models in the field of DRL.

\section{MAIN INSIGHTS}
Below we list the main insights extracted from our study:
\begin{itemize}
\item \textbf{DQN: A Contender for Efficiency:} The study's revelations denote DQN's unexpected superiority in environments with clear, immediate reward structures. DQN's methodical approach to value estimation is well-suited to such scenarios, allowing for a rapid and efficient development of effective strategies. This efficiency is achieved, in part, thanks to the experience replay buffer, which enables DQN to draw from a diverse set of past experiences, preventing over-specialization and promoting a robust policy that generalizes across a multitude of game states.

\item \textbf{Model Adaptability and Hyperparameter Resilience:} DQN's performance exhibits commendable resilience to hyperparameter fluctuations, making it a versatile and forgiving model for practitioners. This adaptability is contrasted with PPO and A2C, which, despite their advanced policy optimization capabilities, show a pronounced sensitivity to hyperparameter settings. This needs a fine-grained tuning process to navigate the trade-offs between exploration and exploitation, especially in environments where the reward pathways are less direct and the strategic demands are higher.

\item \textbf{PPO and A2C for Strategic Exploration:} In environments that reward deep exploration and complex strategy development, PPO and A2C demonstrate their power. Their policy gradient methods become advantageous in more complex state spaces. While these models may require extended periods to converge on highly rewarding strategies, their potential in tasks demanding a sophisticated level of strategic planning is evident.

\item \textbf{Towards a Contextual Model Selection Framework:} The comparative performance of DQN, PPO, and A2C accentuates the necessity for a contextual approach to model selection in DRL. Our findings suggest that while DQN is optimal for tasks demanding quick learning and efficient adaptation, environments characterized by their difficulty and strategic complexity may benefit from the exploratory strengths of PPO and A2C.

\item \textbf{Implications for Practical Application:} The implications of our findings for the practical application of DRL are various. They serve as a guide for practitioners in choosing a model that not only fits the immediate needs of the task but also aligns with the long-term objectives of the learning process. As our understanding of DRL models deepens, it becomes clear that the decision-making framework for selecting a DRL model must be as dynamic and diverse as the models themselves.
\end{itemize}

In conclusion, the study emphasizes the importance of carefully selecting DRL models based on their suitability for the specific task, rather than their perceived complexity. This strategy ensures that the chosen model is both theoretically robust and practically effective, capable of leveraging the unique dynamics of the environment to achieve optimal learning and performance.

\section{Conclusion and Future Work}
This study has detailedly evaluated DQN, PPO, and A2C  within the BreakOut Atari environment, revealing distinct capabilities and performance profiles for each model. DQN, traditionally viewed as a simpler model, has demonstrated its robustness and efficiency in a controlled setting, suggesting that its utility in practical applications remains significant, especially in environments with clear, immediate reward structures. PPO and A2C, while typically favored for complex tasks due to their advanced policy optimization techniques, require careful tuning and strategy development, which can be advantageous in more intricate or less predictable environments. The findings highlight the importance of choosing the right model based on the specific characteristics and requirements of the task environment.

Future research should expand on the comparative analysis of DQN, PPO, and A2C across a wider range of environments, including those with delayed rewards and higher complexity. Further interesting studies could research the impact of additional hyperparameters, network architectures, and reward shaping techniques on the performance of these models.

Lastly, the exploration of DRL applications in real-world scenarios, such as robotics, autonomous vehicles, and financial modeling, where the environment is often unpredictable and complex, will be essential in translating these findings into tangible benefits and advancements in the field of AI.


\clearpage



\clearpage
\appendix

\section{Reproducibility}
To ensure the reproducibility of our results, we have made the code and data used in this study available on GitHub. The repository contains detailed instructions and scripts for replicating the experiments.

\subsection{Repository Overview}
The repository is organized into the following directories, each containing a specific approach to solving BreakOut using different RL algorithms:

\begin{itemize}
  \item \textbf{BreakOut\_sb3\_A2C}: Implementation of the Advantage Actor-Critic (A2C) algorithm using Stable Baselines 3.
  \item \textbf{BreakOut\_sb3\_PPO}: Proximal Policy Optimization (PPO) approach using Stable Baselines 3.
  \item \textbf{BreakOut\_sb3\_DQN}: Implementation of DQN using Stable Baselines 3.
\end{itemize}

\subsection{Installation}
Ensure Python 3.10 is installed along with the following dependencies, which are common across all implementations:
\begin{verbatim}
pip install -r requirements.txt
\end{verbatim}

\subsection{Usage}
1. Clone and set up the repository:
\begin{verbatim}
git clone https://github.com/Neilus03/DRL_comparative_study
cd DRL_comparative_study
pip install -r requirements.txt
\end{verbatim}

2. Navigate to the desired implementation directory, e.g., for PPO:
\begin{verbatim}
cd BreakOut_sb3_PPO
\end{verbatim}

3. Configure the \texttt{config.py} file to adjust the model training, the wandb account, and the saving model options.

4. Execute the \texttt{train.py} to run the model or \texttt{test.py} if you already have a pre-trained model to test.

\subsection{Hardware Requirements}
For reproducing the paper, it is recommended to use a GPU with at least 8 GB of RAM to handle the computational load of training the models.

\end{document}